\documentclass[conference]{IEEEtran}
\IEEEoverridecommandlockouts

\usepackage[hidelinks]{hyperref}
\usepackage[cmex10]{amsmath}
\usepackage{amssymb,amsfonts}
\usepackage{dblfloatfix}
\usepackage{balance}
\usepackage{hyperref}

\usepackage[ruled,vlined]{algorithm2e}
\usepackage{multirow}
\usepackage{graphicx}
\usepackage{array}
\usepackage[table]{xcolor}   
\usepackage{array}           
\usepackage{makecell}

\usepackage{booktabs}
\usepackage{siunitx}
\usepackage[numbers,compress]{natbib}
\usepackage{texnames}
\usepackage{bm,bbm}
\usepackage{orcidlink}
\usepackage{float}

\usepackage{subcaption}
\usepackage{dblfloatfix} 
\usepackage{tabularx}

\usepackage[font=footnotesize,labelfont=bf]{caption}
\usepackage{subcaption}
\usepackage[table]{xcolor}

\title{%
\fontsize{16}{18}\selectfont
EVALUATING TRANSFORMER AND LSTM FRAMEWORKS FOR 
PREDICTION IN UNGAUGED BASINS
}

\author{\IEEEauthorblockN{Taye Akinrele\IEEEauthorrefmark{1},
James Halgren\IEEEauthorrefmark{2},
Noorbakhsh Amiri Golilarz\IEEEauthorrefmark{3}
Sudip Mittal\IEEEauthorrefmark{4},
Shahram Rahimi\IEEEauthorrefmark{5}},
\IEEEauthorblockA{Dept. of Computer Science, The University of Alabama, Tuscaloosa, USA
\\\ Email: \{toakinrele\IEEEauthorrefmark{1}, jshalgren\IEEEauthorrefmark{2}, namirigolilarz\IEEEauthorrefmark{3}, 
smittal1\IEEEauthorrefmark{4}, srahimi\IEEEauthorrefmark{5}\}@ua.edu}
\IEEEauthorblockA{\IEEEauthorrefmark{2}Alabama Water Institute, USA}
}

\begin{document}

\maketitle
\begin{abstract}
Watershed networks exhibit convergent topologies in which multiple tributaries merge into downstream channels, integrating diverse upstream hydrological processes. In ungauged basins, the absence of direct observations increases uncertainty and limits the ability to anticipate extreme events. This study evaluates whether an encoder-only Transformer provides an advantage over an LSTM for upstream streamflow inference under limited hydrologic information, using retrospective simulations from the NOAA National Water Model (NWM). Across both upstream-only and combined configurations, the LSTM showed stronger overall performance than the Transformer model across the two configurations. Incorporating downstream information further boosted performance for all models, increasing median NNSE by more than 60\%. Rather than treating this as a leaderboard-style comparison, we interpret the experiments as a test of architectural inductive bias for hydrologic sequence inference. The results indicate that recurrent memory remains better aligned with this upstream reconstruction task than an encoder-only Transformer, while downstream hydrologic context provides a strong auxiliary constraint that substantially improves prediction skill across architectures.

\end{abstract}


\section{Introduction} \label{sec:intro}

Streamflow measurement using gauging stations and the use of these observations to predict river discharge form the foundation of modern hydrologic forecasting. In recent years, data-driven approaches, particularly deep learning (DL) models, have shown considerable promise in learning complex hydrological relationships directly from data. This has been greatly supported by the emergence of large-sample datasets which provide comprehensive hydrometeorological records across hundreds of catchments, including consistent information on physical attributes, meteorological forcings, and streamflow time series \cite{Coxon2020-jo}, thereby enabling the development of more robust and generalizable models. 

Notably, retrospective datasets such as CAMELS (Catchment Attributes and Meteorology for Large-sample Studies) \cite{Addor2017-wk}, Caravan \cite{Kratzert2023-je}, NWM \cite{opendataNOAANational}, EStreams \cite{Do_Nascimento2024-hd} have become increasingly valuable with the rise of data-intensive machine learning models \cite{Kratzert2023-je}. These developments have encouraged the application of DL models, with the Long Short-Term Memory (LSTM) model achieving notable success in predicting streamflow for ungauged basins. Nearing et al. \cite{Nearing2024-ux} demonstrated that the LSTM network can effectively forecast extreme floods in ungauged settings, while Kratzert et al. \cite{Kratzert2019-sj} showed that LSTMs outperform conceptual models and that sufficient information exists within catchment characteristics to support data-driven modeling under PUB conditions. As a gated recurrent neural network, the LSTM was designed to mitigate the vanishing-gradient limitations of standard RNNs and to better preserve long-range temporal dependencies through its cell-state memory mechanism.

The Transformer architecture has demonstrated outstanding performance across various tasks, including natural language processing, speech recognition, computer vision, and question answering, and has recently been adapted for hydrological modeling applications. Transformer models offer an alternative sequence-learning approach because self-attention can model dependencies across available time steps in parallel during training. Several recent studies have highlighted the effectiveness of Transformers in hydrological forecasting. Yin et al. \cite{Yin2025-rt} proposed the Transformer-XAJ, a process and data-driven model, which achieved strong performance in both regional and ungauged basin predictions. Similarly, Amanambu et al. \cite{Amanambu2022-qm} demonstrated that the Transformer model outperformed LSTM architectures in hydrological drought forecasting across multiple prediction time steps.


Motivated by the success of LSTMs in hydrologic forecasting and the growing use of Transformer architectures in sequence modeling, this study asks whether attention-based models provide a practical advantage over recurrent models for upstream streamflow inference in ungauged basins. 
To guide this investigation, we seek to answer two research questions:
\textbf{RQ1:} Can an LSTM more effectively capture the lagged, state-dependent dynamics of hydrologic response than an encoder-only Transformer under limited upstream information?
\textbf{RQ2:} Does incorporating downstream hydrologic context improve performance across architectures by providing a network-level constraint on upstream reconstruction?

To answer these questions, we:
\begin{itemize}
    \item evaluate recurrent and attention-based architectures under a constrained upstream-only setting using NWM retrospective simulations; 
    \item quantify the effect of adding downstream hydrometeorological context \cite{Ramirez_Molina2025-kg}, and
    \item interpret the resulting performance differences in terms of hydrologic information availability and architectural inductive bias.
\end{itemize}

The goal of this study is to assess which sequence-modeling bias is better suited for upstream streamflow inference under limited information, and how this comparison changes when the downstream hydrologic context is incorporated.


\begin{figure*}
\begin{subfigure}{.5\textwidth}
    \footnotesize
    \captionsetup{font=footnotesize}
    \centering
    \includegraphics[width=0.9\linewidth]{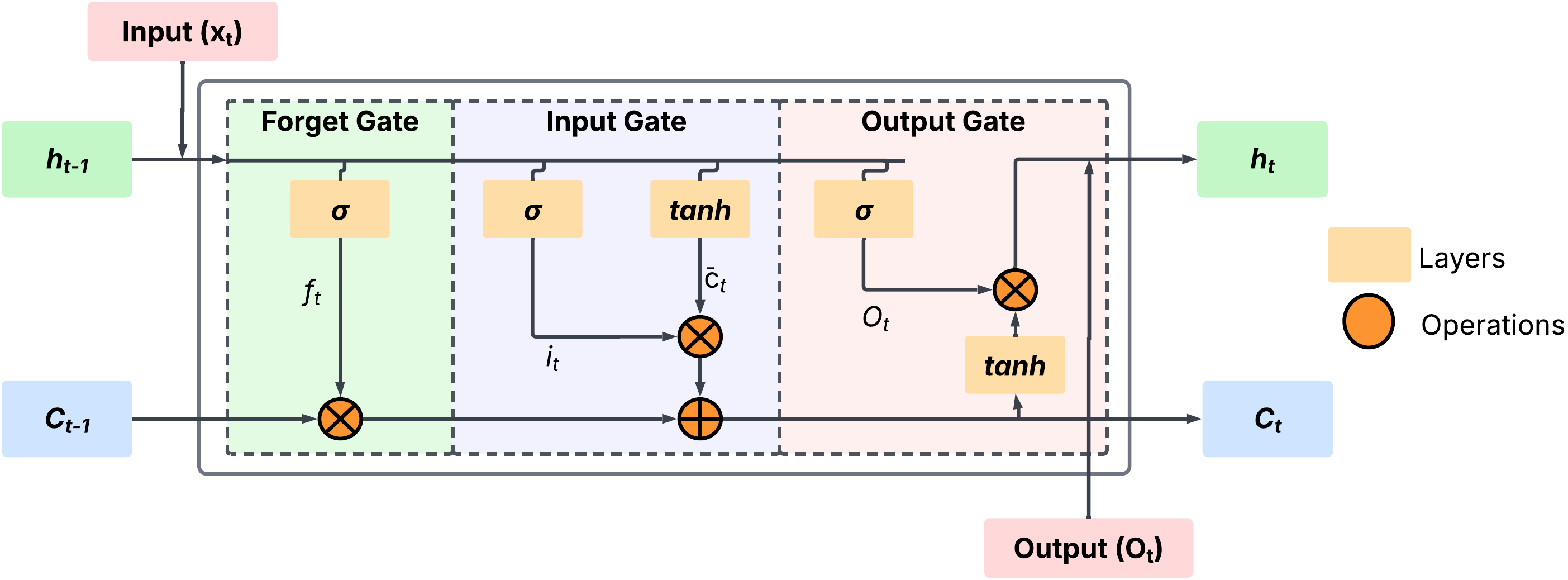}
    \caption{}
\end{subfigure}
\begin{subfigure}{.5\textwidth}
    \footnotesize
    \captionsetup{font=footnotesize}
    \centering
    \includegraphics[width=0.8\linewidth]{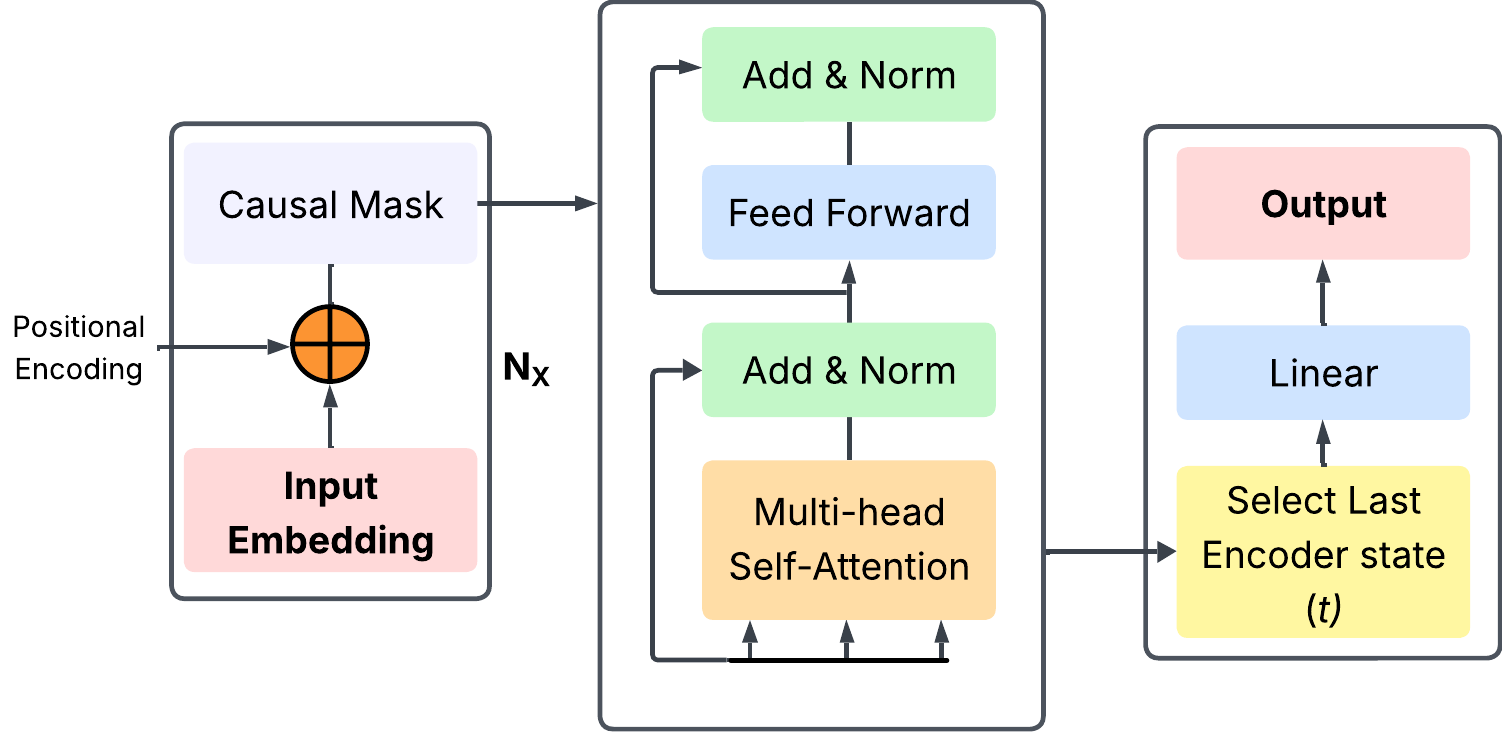}
    \caption{}
\end{subfigure} 
\footnotesize
\captionsetup{font=footnotesize}
\caption{Model architectures used in this study: (a) LSTM model highlighting gate mechanisms ($f_t$, $i_t$, $o_t$) and the interaction between cell state ($C_t$) and hidden state ($h_t$) during streamflow sequence learning (Xu et al., 2020) \cite{Xu2020-kd} (b) Encoder-only Transformer for hydrologic time series modeling, incorporating input embeddings, positional encoding, masked multi-head attention, and feed-forward layers (Vaswani et al., 2017) \cite{NIPS2017_3f5ee243} }
\label{fig:model_architecture}
\end{figure*}

\section{Methodology}\label{sec:methodology}
Given the interdependence between upstream and downstream streamflow, effective modeling requires architectures capable of capturing temporal dependencies and nonlinear hydrologic behavior. This study examines two deep learning approaches, an LSTM-based recurrent model and an encoder-only Transformer, each leveraging different mechanisms for sequence modeling.

The \textbf{LSTM} architecture (Fig.~\ref{fig:model_architecture}a) processes time series inputs sequentially, updating its hidden state through a series of gating operations. The forget, input, and output gates ($f_t$, $i_t$, $o_t$) regulate how information is retained, updated, and exposed at each timestep, enabling the model to learn temporal dependencies through internal recurrence. The cell state ($C_t$) acts as an explicit memory pathway, helping the LSTM maintain information across long sequences. 

In contrast, the encoder \textbf{Transformer} architecture (Fig.~\ref{fig:model_architecture}b) replaces recurrence with parallel self-attention mechanisms. Instead of processing one timestep at a time, the model computes relationships across all timesteps simultaneously, learning how different points in the sequence relate to each other through multi-head self-attention layers. Transformers rely on positional encodings to represent temporal order, since this information is not captured inherently by the architecture. During training, a causal mask is applied to ensure that predictions at each timestep only attend to historical information, making the model suitable for autoregressive hydrologic forecasting. Static catchment attributes and dynamic time series inputs are processed through separate embedding layers before being combined, enabling the model to leverage both time-varying and spatial characteristics of each basin.

\begin{table}[ht]
\centering
\scriptsize
\captionsetup{font=scriptsize}
\caption{Hydrological attributes used in this study, grouped by category (National Water Model Retrospective Dataset v3.0).}
\label{tab:attributes}
\renewcommand{\arraystretch}{1.25}

\begin{tabular}{|m{1.2cm}|m{2.8cm}|m{0.7cm}|m{2.3cm}|}
\hline
\rowcolor{lightgray}\textbf{Category} & \textbf{Attribute} & \textbf{Unit} & \textbf{Description} \\
\hline
\multirow{9}{*}{\shortstack[l]{Dynamic\\Forcings}}
    & APCP\_surface        & mm/s  & Accumulated precipitation \\ \cline{2-4}
    & precip\_rate         & mm/hr  & Precipitation rate \\ \cline{2-4}
    & TMP\_2maboveground   & K        & Near-surface air temperature \\ \cline{2-4}
    & DSWRF\_surface       & W/m$^2$  & Downward shortwave radiation \\ \cline{2-4}
    & DLWRF\_surface       & W/m$^2$   & Downward longwave radiation \\ \cline{2-4}
    & PRES\_surface        & Pa     & Surface pressure \\ \cline{2-4}
    & UGRD\_10maboveground & m/s   & East–west wind at 10m \\ \cline{2-4}
    & VGRD\_10maboveground & m/s   & North–south wind at 10m \\ \cline{2-4}
    & SPFH\_2maboveground  & kg/kg & Specific humidity at 2m \\
\hline
Hydrologic Input & streamflow & m$^{3}$/s & Downstream discharge \\
\hline
\multirow{3}{*}{\shortstack[l]{Catchment\\Attributes}}
    & basin\_length & km       & Length of the basin polygon \\ \cline{2-4}
    & basin\_area   & km$^{2}$ & Area of the basin polygon \\ \cline{2-4}
    & reach\_length & km       & Length of the river reach \\
\hline
Target 
    & streamflow    & m$^3$/s & Upstream discharge (target variable) \\
\hline
\end{tabular}
\end{table}

\section{Results and Discussion}\label{sec: experiments}

\subsection{Data Collection and Integration}
Hourly hydrometeorological data from February 1, 1979, to January 1, 2023 (44 years) were obtained from the NWM v3.0 dataset. USGS gauges were linked to their corresponding reach identifiers using the RouteLink topology for 671 CAMELS basins across CONUS. Upstream reaches were identified from river network connectivity, and basin geometries were merged to define upstream-downstream basin pairs. Dynamic meteorological forcings and static catchment attributes (Table~\ref{tab:attributes}) were included to provide both climatic and physiographic context.

This study uses the simple upstream configuration of $n=1$ reach. For each basin pair, meteorological forcings were spatially averaged over the upstream contributing area, and hourly streamflow was extracted for both downstream and upstream. All variables were stored in NetCDF format with consistent temporal alignment.

\subsection{Data Preprocessing and Training Strategy}
The dataset was processed through a structured pipeline consisting of variable filtering, temporal splitting, normalization, and sequence generation. Filtering removed basins with substantial data gaps; splitting produced training, validation, and test periods; normalization ensured consistent feature scales; and sequence generation converted continuous time series into supervised learning samples.

A 70--15--15 temporal split was used (29 years training, 8 years validation, 8 years testing). Static and dynamic inputs were embedded through fully connected networks (32 units for dynamic inputs and 16 units for static attributes, both with \texttt{tanh} activation and 0.1 dropout) and provided to each model as 256-length input windows. Both the LSTM and Transformer architectures were trained under matched experimental conditions to ensure comparability. The reported configurations were selected as comparable baseline settings under the same data split, optimizer, loss function, input window, and early-stopping rule, rather than through an exhaustive architecture-specific hyperparameter search.

\textbf{LSTM configuration:} consists of a recurrent layer containing 32 hidden units. A forget-gate bias of~3 and an output dropout rate of~0.1 were applied. Static and dynamic inputs were embedded through fully connected layers before being passed to the LSTM, and the final timestep was predicted using a linear regression head.

\textbf{Transformer configuration:} consists of a single Transformer encoder layer with four attention heads, a feedforward dimension of 128, and dropout of 0.1 applied within the encoder and positional encoding layers. Temporal order is represented using sum-based positional encoding, and a causal mask is applied so that each prediction depends only on historical inputs. The final prediction is generated from the last encoder state through a linear regression head.

\textbf{Training setup: } Both models were trained to predict upstream streamflow one hour ahead under two input settings: upstream-only and combined upstream–downstream. Inputs included precipitation, temperature, wind speed, pressure, and radiation. Training used AdamW with a learning rate of $1\times10^{-4}$, batch size 256, gradient clipping of 1, and NSE loss. Early stopping was applied with a patience of 10 epochs, and the best checkpoint was used for testing. The implementation is available \href{https://github.com/Racta-1/hydro-transformer}{here}.

\begin{table}[H]
\centering
\scriptsize
\captionsetup{font=footnotesize}
\setlength{\tabcolsep}{3pt}    
\renewcommand{\arraystretch}{1.25}
\caption{Summary of evaluation metrics used for model performance assessment.}
\label{tab:metrics_summary}
\newcolumntype{C}[1]{>{\centering\arraybackslash}m{#1}}
\begin{tabular}{|m{1.3cm}|m{4.7cm}|C{2.0cm}|}
\hline
\rowcolor{lightgray}
\textbf{Metric} & \textbf{Description} & \textbf{Range, Best Fit} \\
\hline

NNSE & Normalized form of NSE bounded between 0 and 1 for intuitive interpretation. &(0, 1), best: 1 \\
\hline
KGE & Combines correlation, bias, and variability for balanced model evaluation. &($-\infty$, 1), best: 1 \\
\hline
Pearson-$r$ & Measures linear correlation between simulated and observed streamflow. &(-1, 1), best: 1 \\
\hline
RMSE & Quantifies the average magnitude of simulation errors; lower indicates better short-term fit. &(0, $\infty$), best: 0 \\
\hline
\end{tabular}
\end{table}

\subsection{Results \& Performance Analysis} 
To evaluate how model behavior varies with information availability, we trained both the Transformer and LSTM under two distinct input configurations. The upstream-only setting represents a constrained-information setting in which upstream streamflow is predicted solely from local meteorological forcings and static basin attributes. In contrast, the combined setting augments these inputs with downstream meteorological forcings and downstream discharge, introducing a network-informed signal that may reflect the integrated hydrologic response of connected reaches. This design allows us to test not only whether additional inputs improve accuracy, but also how each architecture responds to a richer hydrologic context.

\begin{figure*}
\begin{subfigure}{.33\textwidth}
    \footnotesize
    \captionsetup{font=scriptsize}
    \centering
    \includegraphics[width=\linewidth]{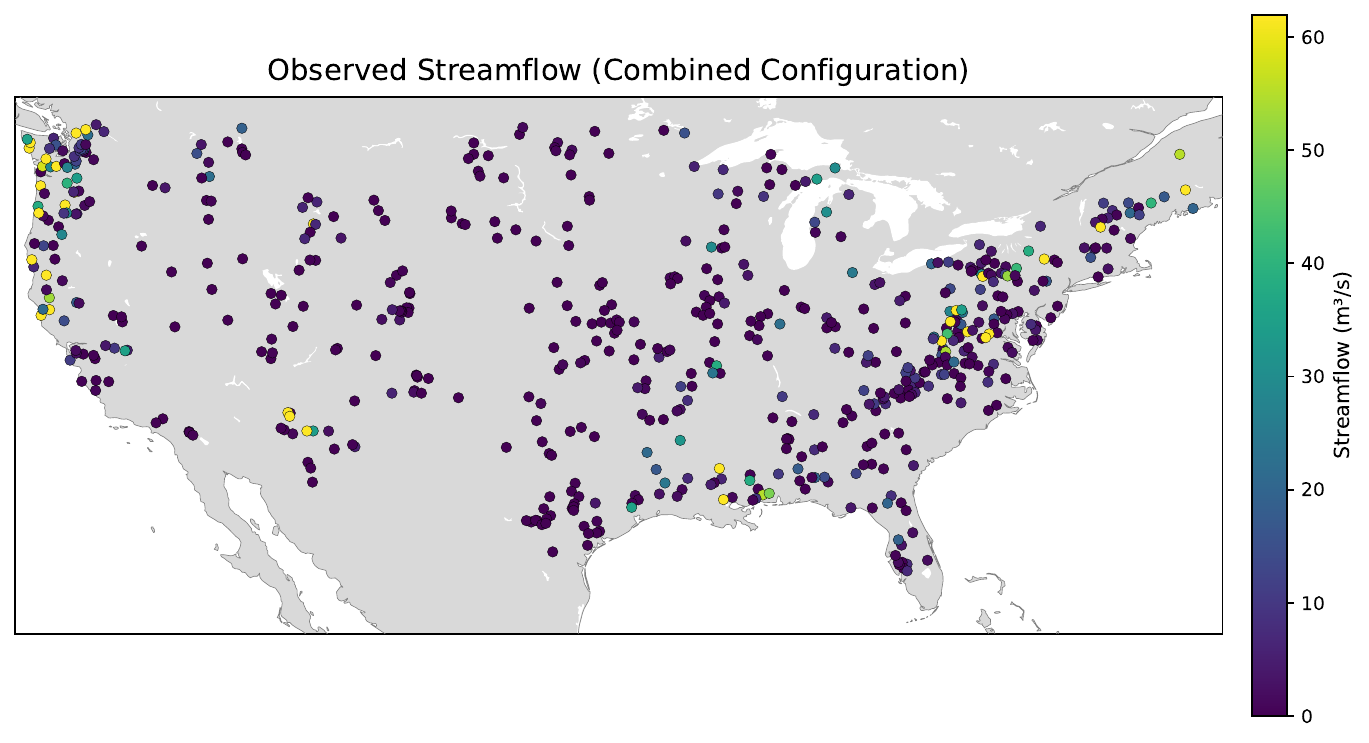}
    \caption{}
\end{subfigure}
\begin{subfigure}{.33\textwidth}
    \footnotesize
    \captionsetup{font=scriptsize}
    \centering
    \includegraphics[width=\linewidth]{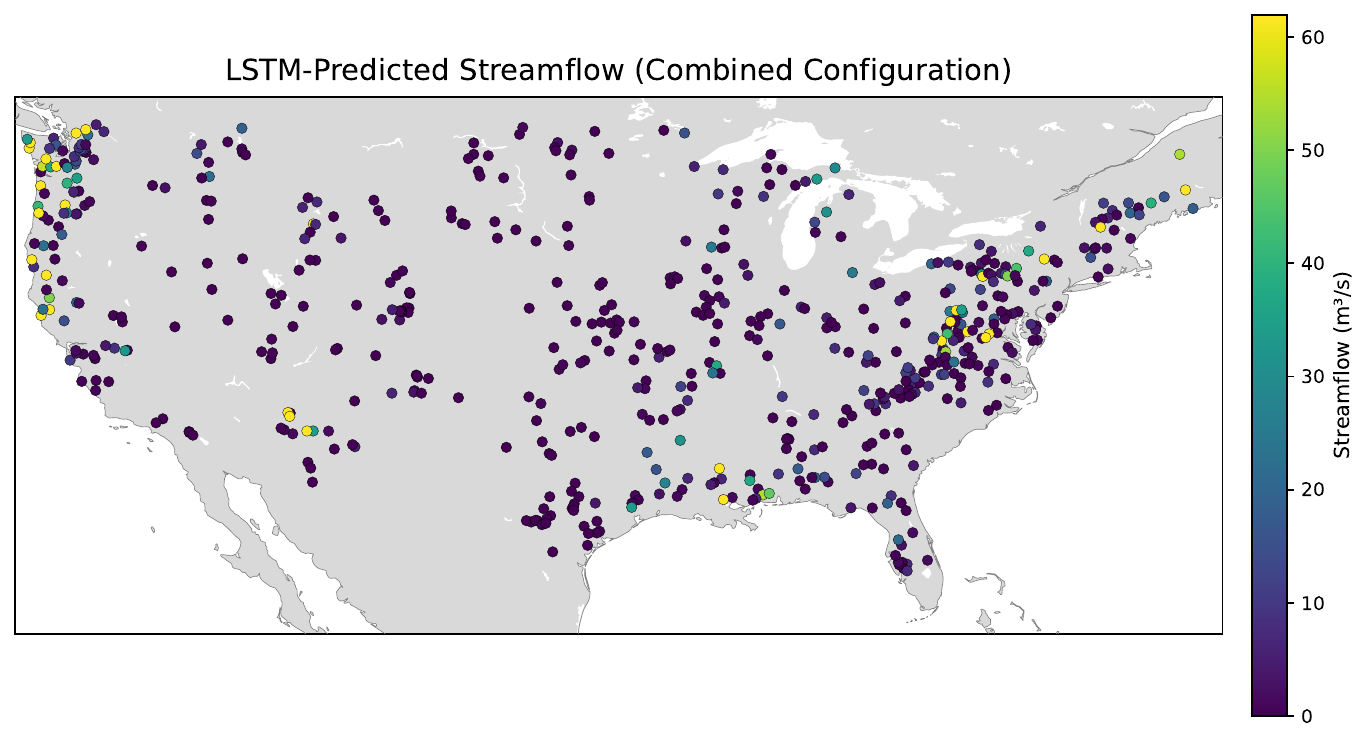}
    \caption{}
\end{subfigure}
\begin{subfigure}{.33\textwidth}
    \footnotesize
    \captionsetup{font=scriptsize}
    \centering
    \includegraphics[width=\linewidth]{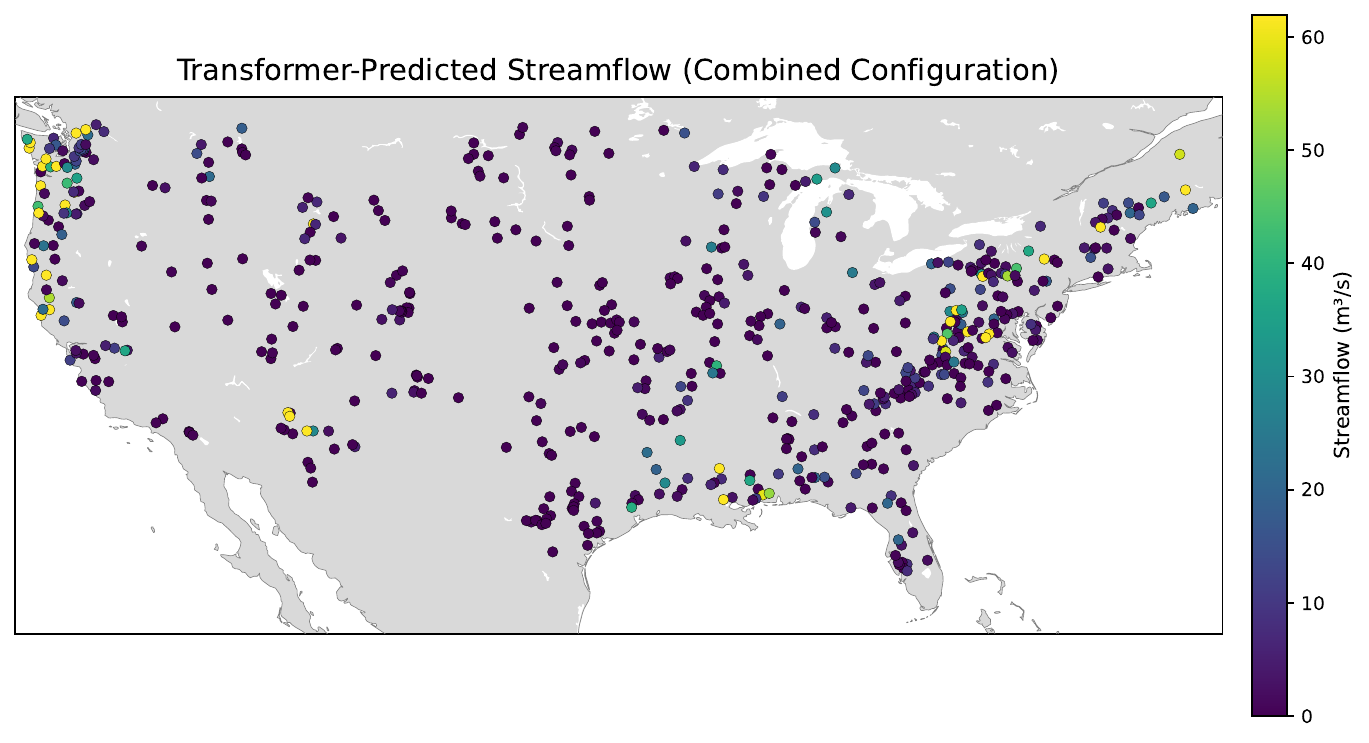}
    \caption{}
\end{subfigure}
\footnotesize
\captionsetup{font=footnotesize}
\caption {Spatial comparison of observed and model-predicted streamflow at upstream gages under the combined configuration across the CONUS: \textbf{(a)} observed streamflow, \textbf{(b)} LSTM predicted streamflow, and \textbf{(c)} Encoder-only Transformer predicted streamflow. Colors indicate streamflow magnitude in m³/s using a common scale across all panels.}
\label{fig:spatial_comparison}
\end{figure*}


\begin{figure*}
\begin{subfigure}{.33\textwidth}
    \footnotesize
    \captionsetup{font=scriptsize}
    \centering
    \includegraphics[width=\linewidth]{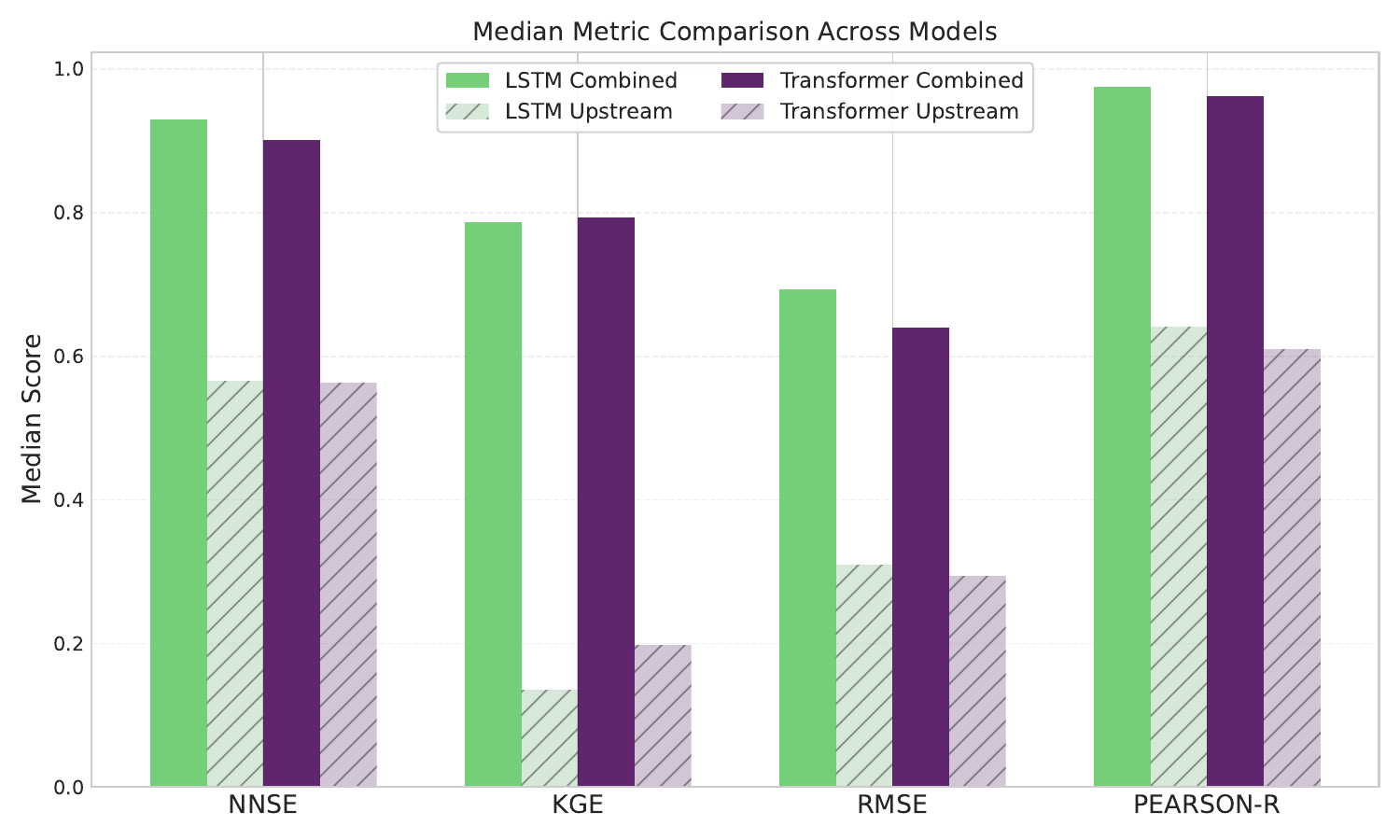}
    \caption{}
\end{subfigure}
\begin{subfigure}{.33\textwidth}
    \footnotesize
    \captionsetup{font=scriptsize}
    \centering
    \includegraphics[width=\linewidth]{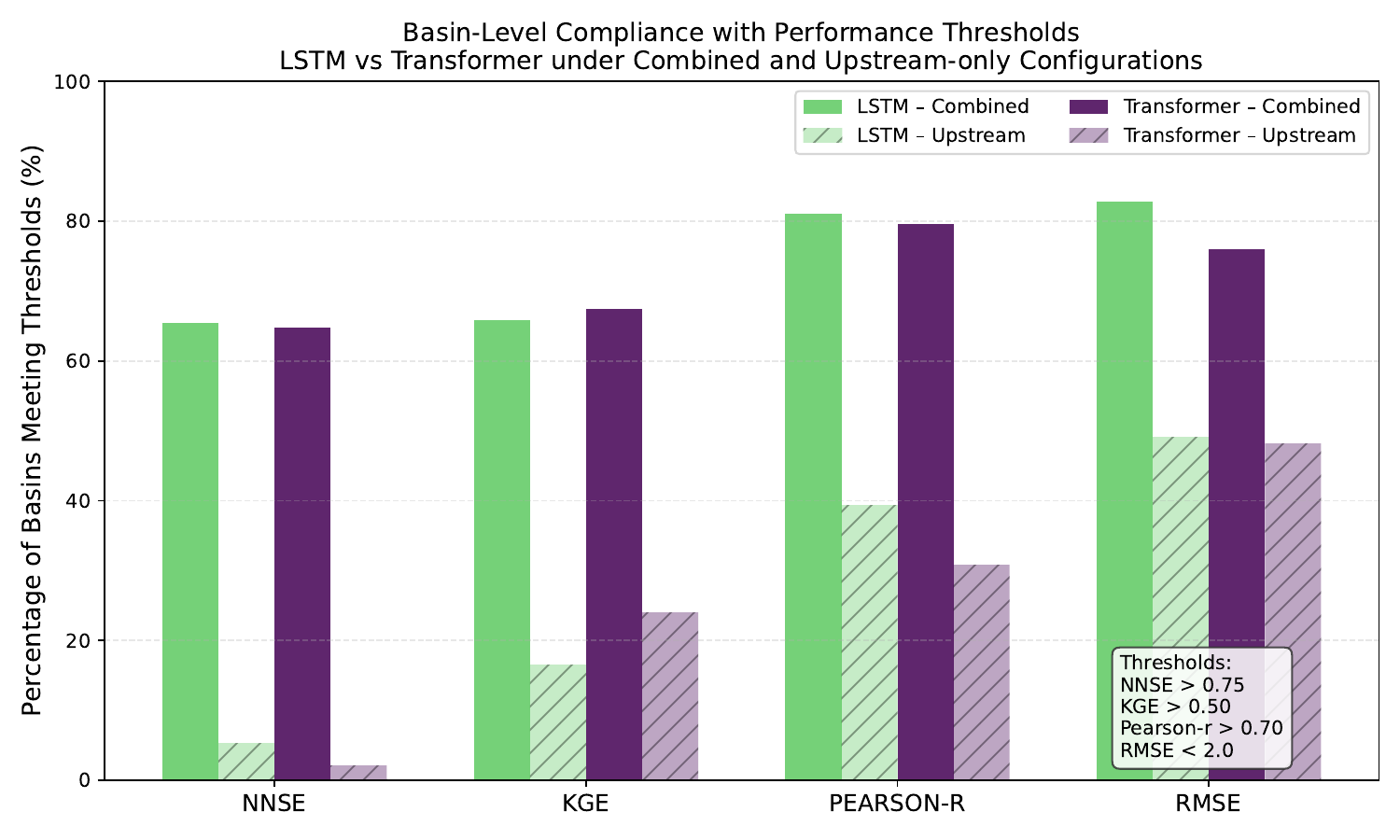}
    \caption{}
\end{subfigure}
\begin{subfigure}{.33\textwidth}
    \footnotesize
    \captionsetup{font=scriptsize}
    \centering
    \includegraphics[width=\linewidth]{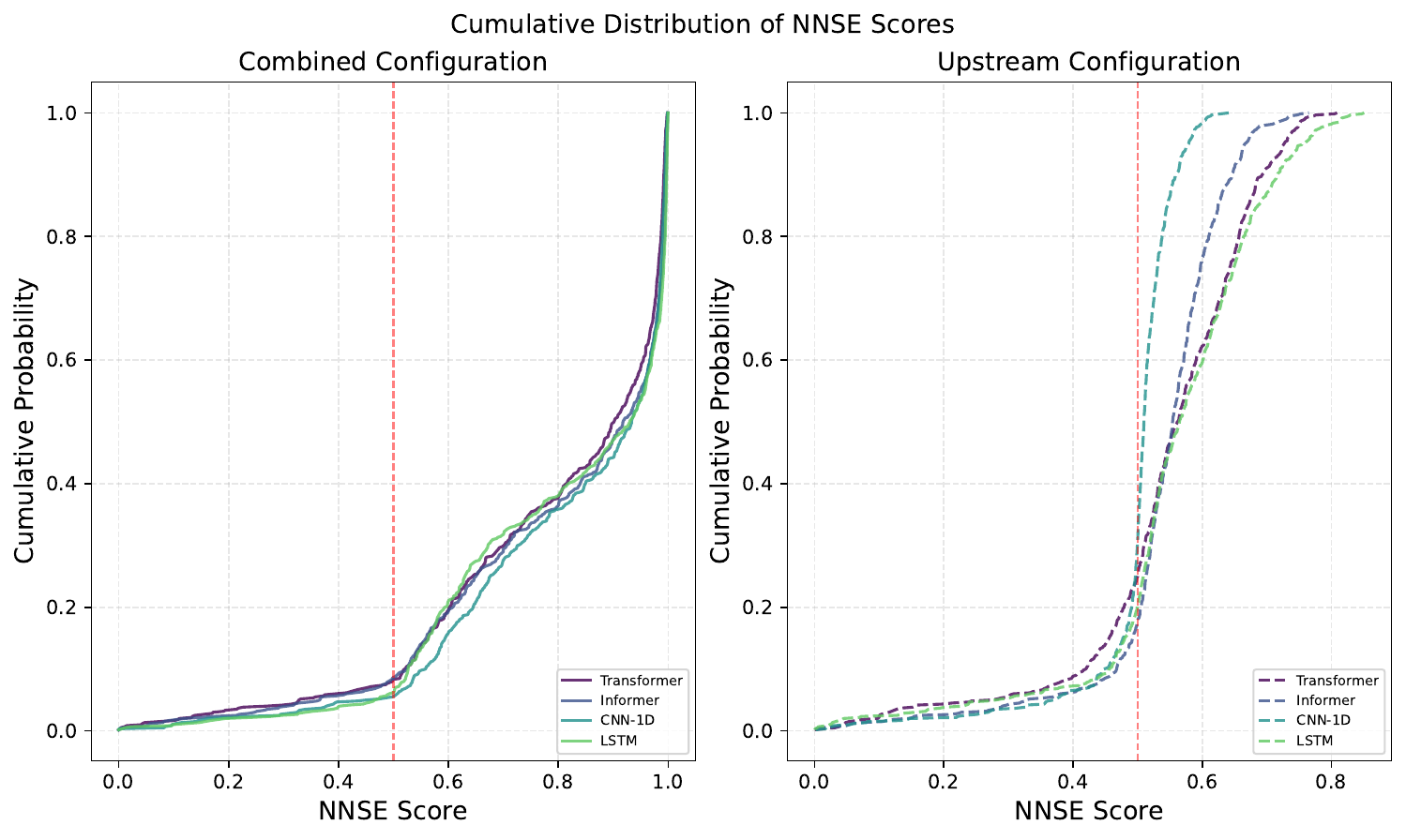}
    \caption{}
\end{subfigure}
\footnotesize
\captionsetup{font=footnotesize}
\caption {Comparison of model performance under the Combined and Upstream configurations. 
\textbf{(a)} Median basin-level performance across NNSE, KGE, RMSE, and Pearson-r for LSTM and Transformer models, highlighting improvements from incorporating downstream information.
\textbf{(b)} Percentage of basins meeting predefined performance thresholds (NNSE $>$ 0.75, KGE $>$ 0.50, Pearson-r $>$ 0.70, RMSE $<$ 2.0).
\textbf{(c)} Cumulative distribution functions (CDFs) of NNSE scores across LSTM, Transformer, Informer, and CNN-1D models (Combined (left) and Upstream (right) setups). \textit{\textbf{Note that lower RMSE values indicate better performance, unlike the other metrics.}}}
\label{fig:comparison_models}
\end{figure*}

As shown in Table~\ref{tab:up_vs_combined}, the combined configuration consistently outperformed the upstream-only setup for both Transformer and LSTM models. For example, median NNSE for the Transformer model increases from 0.56 to 0.90, and for the LSTM from 0.56 to 0.93, demonstrating substantial improvements in predictive accuracy. A similar trend is observed for KGE, Pearson-$r$, and RMSE, indicating that the combined inputs enhance robustness and reduce error magnitudes across basins. The percentage of basins achieving NNSE $> 0.5$ also increases significantly in the combined setting, for the Transformer from 75.21\% to 91.68\%, and for the LSTM from 79.87\% to 93.84\%, highlighting more reliable performance across a larger fraction of the domain (see Fig.~\ref{fig:comparison_models}b). 

This experiment shows how much predictive skill can be recovered from network-level hydrologic context and whether that improvement varies by architecture. Both models benefit substantially from downstream information, but the LSTM shows stronger overall performance across the two configurations. This suggests that richer hydrologic context improves inference, but does not diminish the importance of an architecture well suited to sequential runoff dynamics.

\begin{table}
\centering
\scriptsize
\captionsetup{font=footnotesize}
\setlength{\tabcolsep}{3pt}
\renewcommand{\arraystretch}{1.25}
\caption{Performance comparison of the encoder-only Transformer and LSTM models for both configurations across all basins.}
\label{tab:up_vs_combined}
\newcolumntype{C}[1]{>{\centering\arraybackslash}m{#1}}
\begin{tabular}{m{2.8cm} C{1.2cm} C{1.2cm} C{1.2cm} C{1.2cm}}
\toprule
\multirow{2}{*}{\textbf{Metric}} 
& \multicolumn{2}{c}{\textbf{Transformer}} 
& \multicolumn{2}{c}{\textbf{LSTM}} \\
& Upstream & Combined
& Upstream & Combined \\
\midrule
NNSE        & 0.56 & 0.90 & 0.56 & \textbf{0.93} \\
KGE         & 0.20 & \textbf{0.80} & 0.14 & 0.79 \\
Pearson-$r$ & 0.61 & 0.96 & 0.64 & \textbf{0.98} \\
RMSE        & 2.41 & 0.56 & 2.24 & \textbf{0.44} \\
\midrule
\textbf{\% Basins NNSE $> 0.5$} 
& 75.21\% & 91.68\% & \textbf{79.87}\% & \textbf{93.84}\% \\
\midrule
\textbf{Basin Count} & \multicolumn{4}{c}{601} \\
\bottomrule
\end{tabular}
\end{table}

Overall, the LSTM shows modest but consistent advantages over the encoder-only Transformer across both configurations, with slightly higher NNSE and Pearson-$r$ and lower RMSE, while KGE remains comparable as seen in Fig.~\ref{fig:comparison_models}a. The difference in median NNSE in the combined setting (0.93 vs. 0.90) is small, indicating that both models achieve similar overall predictive skill.

These results suggest that downstream information is the primary driver of performance gains, while the LSTM provides a modest relative benefit for streamflow reconstruction. One plausible explanation is that upstream inference is governed by temporally accumulated and state-dependent hydrologic processes. The LSTM’s sequential state updates align naturally with these dynamics, whereas the encoder-only Transformer must infer temporal structure through attention and positional encoding alone, which appears slightly less effective in this setting.


To assess spatial consistency, we compare observed and predicted streamflow across CONUS. As shown in Fig.~\ref{fig:spatial_comparison}, both the LSTM and the encoder-only Transformer capture similar large-scale spatial patterns under the combined configuration.

\subsection{Performance Comparison of Other Models}
In Fig.~\ref{fig:comparison_models}c, we extend the comparison to CNN-1D \cite{Van2020-kv} and Informer \cite{Zhou_Zhang_Peng_Zhang_Li_Xiong_Zhang_2021}. Across architectures, NNSE distributions become more similar when downstream hydrologic information is included, with most basin-level scores concentrated between 0.8 and 1.0, indicating reduced performance differences between models. A plausible explanation is that downstream observations act as a strong network-level constraint, reducing uncertainty and limiting the influence of model-specific inductive biases. In contrast, removing downstream context leads to a general decline in performance, with CNN-1D showing greater variability and lower median NNSE, while Informer, Transformer, and LSTM remain comparatively more stable. This suggests that architectures capable of capturing long-range dependencies or maintaining temporal state are better able to compensate when hydrologic context is limited. Overall, downstream information improves performance and narrows the gap between model classes.

\section{Conclusion and Future Work}\label{sec:conclusion}
In this study, we compared LSTM and encoder-only Transformer architectures for streamflow prediction using NWM data across CAMELS basins under two inference settings: an upstream-only configuration reflecting PUB-style constraints, and a network-informed configuration that incorporates downstream hydrologic context. The latter is not a strict PUB setting, but represents a scenario where additional network-level information is available.

Results show that incorporating downstream information consistently improves performance across architectures, indicating that network-level context plays a dominant role in capturing spatial hydrologic dependencies. While both models benefit substantially, the LSTM achieves modestly higher accuracy and more stable basin-wise performance, suggesting that recurrent and attention-based architectures respond differently to information scarcity and network augmentation.

Future work will extend this framework to larger upstream–downstream networks and to observed USGS discharge data, enabling broader validation under more realistic hydrologic conditions.


\section{Acknowledgments}

This research was supported by the Cooperative Institute for Research to Operations in Hydrology (CIROH) with funding under award NA22NWS4320003 from the NOAA Cooperative Institute Program. The statements, findings, conclusions, and recommendations are those of the author(s) and do not necessarily reflect the opinions of NOAA.


\clearpage
\balance
\bibliographystyle{IEEEtran}
\footnotesize
\bibliography{references}
\end{document}